\documentclass[iicol,sn-basic,Numbered]{sn-jnl}

\usepackage{graphicx}%
\usepackage{multirow}%
\usepackage{amsmath,amssymb,amsfonts}%
\usepackage{amsthm}%
\usepackage{mathrsfs}%
\usepackage[title]{appendix}%
\usepackage{xcolor}%
\usepackage{textcomp}%
\usepackage{manyfoot}%
\usepackage{booktabs}%
\usepackage{algorithm}%
\usepackage{algorithmicx}%
\usepackage{algpseudocode}%
\usepackage{listings}
\usepackage{array}
\usepackage{makecell}
\usepackage{color,soul}

\usepackage{subcaption}
\usepackage{multirow}

\newcommand{\STAB}[1]{\begin{tabular}{@{}c@{}}#1\end{tabular}}

\begin{document}

\title[L-VAE]{L-VAE: Variational Auto-Encoder with Learnable Beta for Disentangled Representation}

\author*[1]{\fnm{Hazal} \sur{Mogultay Ozcan}}\email{mogultay@metu.edu.tr}

\author[1]{\fnm{Sinan} \sur{Kalkan}}\email{skalkan@metu.edu.tr}

\author[1]{\fnm{Fatos T.} \sur{Yarman-Vural}}\email{yarman@metu.edu.tr}

\affil[1]{\orgdiv{Computer Engineering}, \orgname{Middle East Technical Universty}, \orgaddress{ \city{Ankara}, \postcode{06800}, \country{Turkey}}}

\abstract{In this paper, we propose a novel model called Learnable VAE (L-VAE), which learns a disentangled representation together with the hyperparameters of the cost function. L-VAE can be considered as an extension of  $\beta$-VAE, wherein the hyperparameter, $\beta$, is empirically adjusted. L-VAE mitigates the limitations of $\beta$-VAE by learning the relative weights of the terms in the loss function to control the dynamic trade-off between disentanglement and reconstruction losses. In the proposed  model, the weight of the loss terms and the parameters of the model architecture  are learned concurrently. An additional regularization term is added to the loss function  to prevent bias towards either reconstruction or disentanglement losses. Experimental analyses  show that the proposed L-VAE finds an effective balance between reconstruction fidelity and disentangling the latent dimensions. Comparisons of the proposed L-VAE  against $\beta$-VAE, VAE, ControlVAE, DynamicVAE, and $\sigma$-VAE on datasets, such as dSprites, MPI3D-complex, Falcor3D, and Isaac3D   reveals that L-VAE consistently provides the best or the second best performances measured by  a set of disentanglement metrics. Moreover,  qualitative experiments on CelebA dataset,  confirm the success of the L-VAE  model for disentangling the facial attributes.}

\maketitle

\section{Introduction}

Deep learning architectures inherently possess limitations concerning their capacity for  generalization, explainability, and interpretability \cite{zhang2021understanding,pouyanfar2018survey,bengio2013representation,higgins2016beta}. A promising approach to reduce these limitations is to identify the independent factors of the data generation process that can  represent the implicit properties of the data, such as rotation, translation, shape, or shadow \cite{bengio2013representation,higgins2016beta}. This approach, coined as disentanglement, facilitates generalization, explainability, and interpretability in terms of the identified essential properties \cite{higgins2018definition,locatello2020weakly, Do2020}.

Variational Auto-Encoders (VAEs) \cite{kingma2013auto} are one of the pioneering models to learn disentangled representations. Rather than learning a representation with entangled properties, zero mean unit variance Gaussian priors are enforced on each dimension of the latent properties. For this purpose, the Kullback-Leibler divergence between the priors and the learnt distribution ($\mathcal{D}_{KL}$) is used alongside a reconstruction loss ($\mathcal{L}_\textrm{R}$) to enforce disentanglement in the learnt representation, 
\begin{equation}
    \mathcal{L} = \mathcal{L}_\textrm{R} +  \mathcal{D}_{KL},
\end{equation}
which joins the two incompatible terms with different orders of magnitude and difficulty, which may result in an imbalance in the optimization of the overall loss function. In order to reduce this imbalance problem, the KL divergence term is weighted by a hyperparameter $\beta$ in $\beta$-VAE \cite{higgins2016beta, burgess2018understanding}, which accentuates the importance of the KL divergence with the hope of learning more disentangled representations compared to VAEs. 

Despite its promises, $\beta$-VAE exhibits certain shortcomings for learning disentangled representations. Firstly, the performances of $\beta$-VAE's are not consistently  robust against different values of $\beta$ \cite{locatello2019challenging}. Furthermore, finding an optimal $\beta$ value via empirical methods requires computationally expensive and exhaustive search methods \cite{rybkin2021simple}. Secondly, even when a near-optimal $\beta$ parameter is empirically  caught, minimizing the loss function and the KL divergence with a fixed $\beta$ parameter does not necessarily result in a minimum reconstruction loss with a maximum degree of disentanglement. In most cases, it is observed that $\beta$-VAE increases the reconstruction loss for the sake of better disentanglement \cite{chen2018isolating}, which may result in poor representation of the data. In other words, a fixed $\beta$ parameter makes it difficult to discover the disentangled factors of variations providing minimum reconstruction loss. These challenges indicate the need for better approaches to improve the  disentanglement capabilities of $\beta$-VAE.

In this study, we propose a model called Learnable Variational Auto-Encoder (L-VAE), which learns the  parameter $\beta$ that minimizes the reconstruction loss ($\mathcal{L}_R$) and maximizes disentanglement (through $\mathcal{D}_{KL}$). In this sense, L-VAE can be considered an extended version of $\beta$-VAE. L-VAE, with the learnable $\beta$, learns the trade-off between the reconstruction loss ($\mathcal{L}_R$) and the amount of disentanglement through the  KL divergence term ($\mathcal{D}_{KL}$). This eliminates the burden of optimizing the hyperparameter $\beta$.  

Our main contributions can be summarized as follows:

\begin{itemize}
    \item We conduct an analysis on $\beta$-VAE and highlight several critical observations regarding the $\beta$ hyperparameter. For example, we observe that $\beta$-VAE is highly sensitive to the $\beta$ hyperparameter and that $\beta<1$ can surprisingly provide better disentanglement. 

    \item Motivated by the challenges and the issues associated with the $\beta$ hyperparameter, we propose an auto-differentiation compatible method for learning the $\beta$ hyperparameter in a hassle-free manner.
    
    \item We perform comprehensive evaluations on five different datasets (dSprites, MPI3D-complex, Falcor3D, Isaac3D, and CelebA) and compare L-VAE with VAE and $\beta$-VAE as well as three state-of-the-art methods (ControlVAE \cite{shao2020controlvae, shao2021controlvae}, DynamicVAE \cite{shao2022rethinking}, and $\sigma$-VAE \cite{rybkin2021simple}) which also aim to dynamically tune the $\beta$ parameter in $\beta$-VAE. We show that L-VAE consistently provides the best or second-best performance in terms of widely used disentanglement measures without empirically tuning the $\beta$ hyperparameter. 
\end{itemize}

\section{Disentanglement in Machine Learning}
\label{sec:relatedwork}

Over the last decade, disentangled representation learning received significant attention in the machine learning community. It has been utilized in a wide range of problems, including e.g., facial image analysis \cite{liu2018exploring}, image dehazing \cite{dong2020fd}, face hallucination \cite{duan2020cross}, video frame generation \cite{hsieh2018learning, denton2017unsupervised,zhu2020s3vae},  identity learning \cite{liu2018exploring, nitzan2020face}, image-to-image translation \cite{lee2020drit++}, and face forgery detection \cite{fu2023multi}. In addition to Variational Auto-Encoder-based approaches (to be covered in the next section), researchers have proposed several approaches, such as the ones based on Generative Adversarial Networks \cite{chen2016infogan, Nguyen-Phuoc2019, karras2019style,karras2020analyzing},
and causality \cite{scholkopf2021toward, yang2021causalvae}. Disentanglement is generally studied in unsupervised learning problems, without using any labels for the factors of variations. However, it is possible to address disentanglement in a supervised framework (see, e.g., \cite{tran2017disentangled}).

\subsection{Variational Auto-Encoders and its Variations}

Due to its relative simplicity and disentangling capability,  Variational Auto-Encoders \cite{kingma2013auto} are widely used for disentangled
representation learning. It has been shown that VAE models disentangle the representation space to a certain degree, making them an appealing approach for improving the generalization capacity and interpretability of the model in an unsupervised setting \cite{higgins2016beta}.  $\beta$-VAE encourages  disentanglement by introducing the weight $\beta$ on the KL term. This hyperparameter emphasizes the importance of disentanglement relative to the reconstruction loss. However, this approach introduces a new hyperparameter to be empirically optimized in a large search space. Furthermore, optimization of the overall loss function may result in an increased reconstruction loss.  

Many studies explored the generalization properties of VAEs and proposed extensions. For example,
in order to regain the model's reconstruction abilities, Burgess et al. proposed a control mechanism on the capacity of the model \cite{burgess2018understanding}. They argue that, by gradually increasing the  capacity of the model from zero, they can increase the disentanglement ability, while conserving the reconstruction loss. 

Kim et al. \cite{kim2018disentangling} also argued that minimizing the loss function of the $\beta$-VAE decreases the reconstruction quality.
In order to balance the reconstruction loss and KL divergence, they proposed a new model called Factor-VAE, in which they introduced a discriminator to the same architecture to elaborate the trade-off between disentanglement and reconstruction. The overall loss function increases the independence among latent dimensions and does not affect the mutual information.

Chen et al. \cite{chen2018isolating} investigated the source of success in $\beta$-VAE and decomposed the ELBO loss term into three parts,  namely, index code mutual information, total correlation, and dimension-wise KL divergence. These terms correspond to the mutual information between data and latent code, dependence among variables, and Kullback-Leibler divergence of the terms from the priors, respectively. They argue that the success of $\beta$-VAE in terms of disentanglement stems from the total correlation. Then, they proposed the model $\beta$-TCVAE, an extension to $\beta$-VAE, where they weighted each of the three terms. The optimized function is the same as \cite{kim2018disentangling}; however, they estimated the TC term with a different method. 

A good resource that compares different VAE models is the study by Locatello et al.
\cite{locatello2019challenging}. They analyzed the performances of the aforementioned explicit models in a compelling scenario. They executed 12K models and provided comparative results on six different metrics. Their findings emphasize that hyperparameter selection is crucial in terms of disentanglement.

\subsection{Automatically Tuning $\beta$ in $\beta$-VAE} \label{sec:auto-tuning}
Several studies  explored automatic hyperparameter tuning in VAE models, specifically balancing the reconstruction loss and KL divergence terms in the loss function \cite{shao2020controlvae, shao2021controlvae, shao2022rethinking}. Shao et al. proposed ControlVAE algorithm for this specific purpose \cite{shao2020controlvae, shao2021controlvae}. They designed a Proportional–Integral Controller (PI Controller), a variation of PID controller \cite{pidcontrol}, to automatically adjust the $\beta$ of $\beta$-VAE algorithm. They first set a desired KL value, and at each iteration, they compute the error between the desired KL and its current value and apply a correction to reduce the error. The correction is conducted via the output of the PI controller, i.e., the $\beta$ parameter. They further improved this model and proposed DynamicVAE \cite{shao2022rethinking}. DynamicVAE uses a similar PI Controller, however, it reduces $\beta$ iteratively from a large value rather than increasing it, providing a smoother change on the KL term. Although ControlVAE and DynamicVAE can efficiently learn $\beta$, they introduce several new parameters to optimize, such as the desired value of the KL divergence, the initial value of $\beta$, and the constants of the PID controller. Moreover, Rybkin et al. proposed $\sigma$-VAE, which learns the variance of the decoder with network parameters \cite{rybkin2021simple}. Although they did not analyze $\sigma$-VAE as a disentanglement model and focused on the reconstruction of the model, their work is similar to ours since they learn the relative weight of the reconstruction term.

\subsection{Comparative Summary} 

Most of the aforementioned studies rely on tuning the $\beta$ hyperparameter empirically through extensive experiments, with the exception of the recent ControlVAE \cite{shao2020controlvae,shao2021controlvae}, DynamicVAE \cite{shao2022rethinking} and $\sigma$-VAE \cite{rybkin2021simple} models. Although they show promising results, ControlVAE and DynamicVAE introduce additional hyperparameters, such as, the desired divergence value, or parameters of the PID control algorithm. On the other hand,  $\sigma$-VAE requires soft clipping on the learned weights and implicitly sets a hyperparameter for the learned weight. These additional hyperparameters further complicate a learning disentangled representation. In contrast, we propose a relatively simple and hassle-free solution that does not require  tuning the hyperparameters of the cost function.

In the following sections, we start by  defining the concept of disentanglement. Subsequently, we provide a brief description and analysis of  our observations pertaining to  $\beta$-VAE. Next, we introduce our  Learnable Variational Autoencoder (L-VAE) model. Finally, we report our experiments for analyzing the L-VAE model, comparing it  with the state of the art disentangled learning models.

\section{Definition(s) of Disentanglement}

There are various definitions for disentanglement in representation learning \cite{bengio2013representation,higgins2016beta,higgins2018definition,burgess2018understanding,chen2018isolating,locatello2019challenging}.
A  mathematically tractable approach to define disentanglement is based on the assumption that the data consists of a set of hidden properties shared across the categories. These hidden properties are one of the major causes of the variations among the objects of the same category. Then, disentanglement is defined as the separation of these properties embedded in the observations while learning a  representation function. In this study, we adopt the latter definition and concentrate on the computational aspect of disentanglement. In this definition, observations are assumed to be generated by a set of independent Gaussian sources. The goal of disentanglement is to learn a representation, where each  Gaussian distribution, generated by a source is modeled compactly in a separate subspace of the learnt representation. This approach allows an interpretation of the distribution of the representation in terms of the distribution of the hidden properties.

Formally, consider a dataset $\mathcal{D= \{X, W,Y\}}$, where
$\mathbf{x}_i \in \mathcal{X } \in 	\mathbb{R}^{M\times N}$ denotes an $M\times N$-dimensional image sample $i$, generated by a mixture of $K$ \textit{conditionally independent} and \textit{unobserved} factors of variation 
$\mathbf{w}_i \in \mathcal{W} \in 	\mathbb{R}^K$. Hence, in an unsupervised setting, $\mathbf{x}_i$ can be simulated using its source factors of variation, expressed as, 
$$\mathbf{x}_i \sim Sim(\mathbf{w}_i).$$
In the supervised setting, the dataset can also include the labels of $\mathbf{w}_i$ as $\mathbf{y}_i \in \mathcal{Y} \in \mathbb{R}^K$. 

Disentanglement is then defined as the problem of learning an $L$-dimensional representation $\mathbf{z}_i\in \mathbb{R}^L$ for image sample $\mathbf{x}_i$, with $$p(\mathbf{x}_i|\mathbf{z}_i) \sim Sim(\mathbf{w}_i).$$
In this representation, for each dimension $j$ of the hidden variable vector,
$$\mathbf{z}_i = [\mathrm{z}_i^j],$$
the algorithm  learns a \textit{single} source of variation $\mathrm{w}_i^j$. Note that in theory, $L$ should be selected as $L \geq K$ in order to capture the discriminative information about all of the hidden properties.

\begin{figure*}[!ht]
    \centering
    \begin{subfigure}{0.49\linewidth}  
        \includegraphics[width=\textwidth]{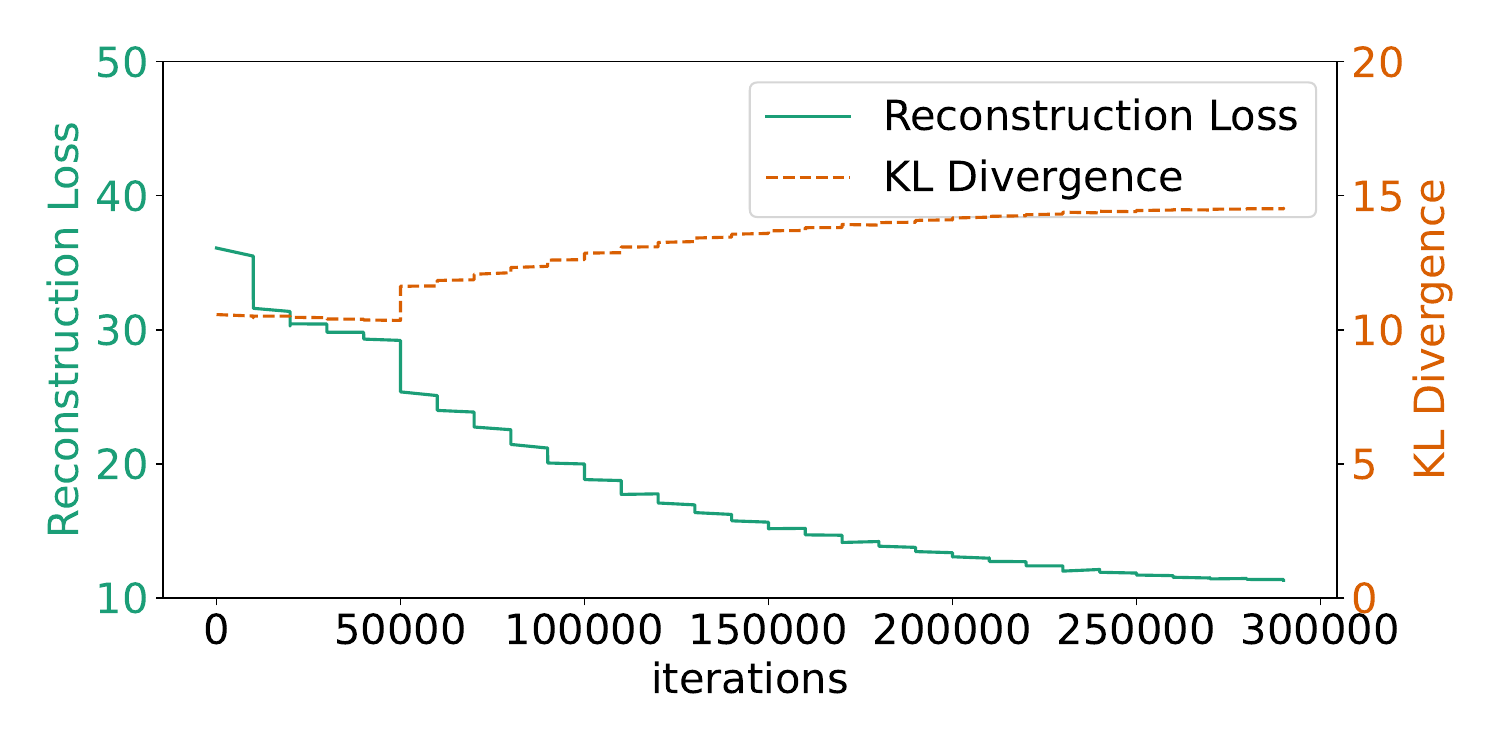}
         \caption{ }
          \label{fig:vae-reconvskl}
    \end{subfigure}
    \begin{subfigure}{0.49\linewidth}
        \includegraphics[width=\textwidth]{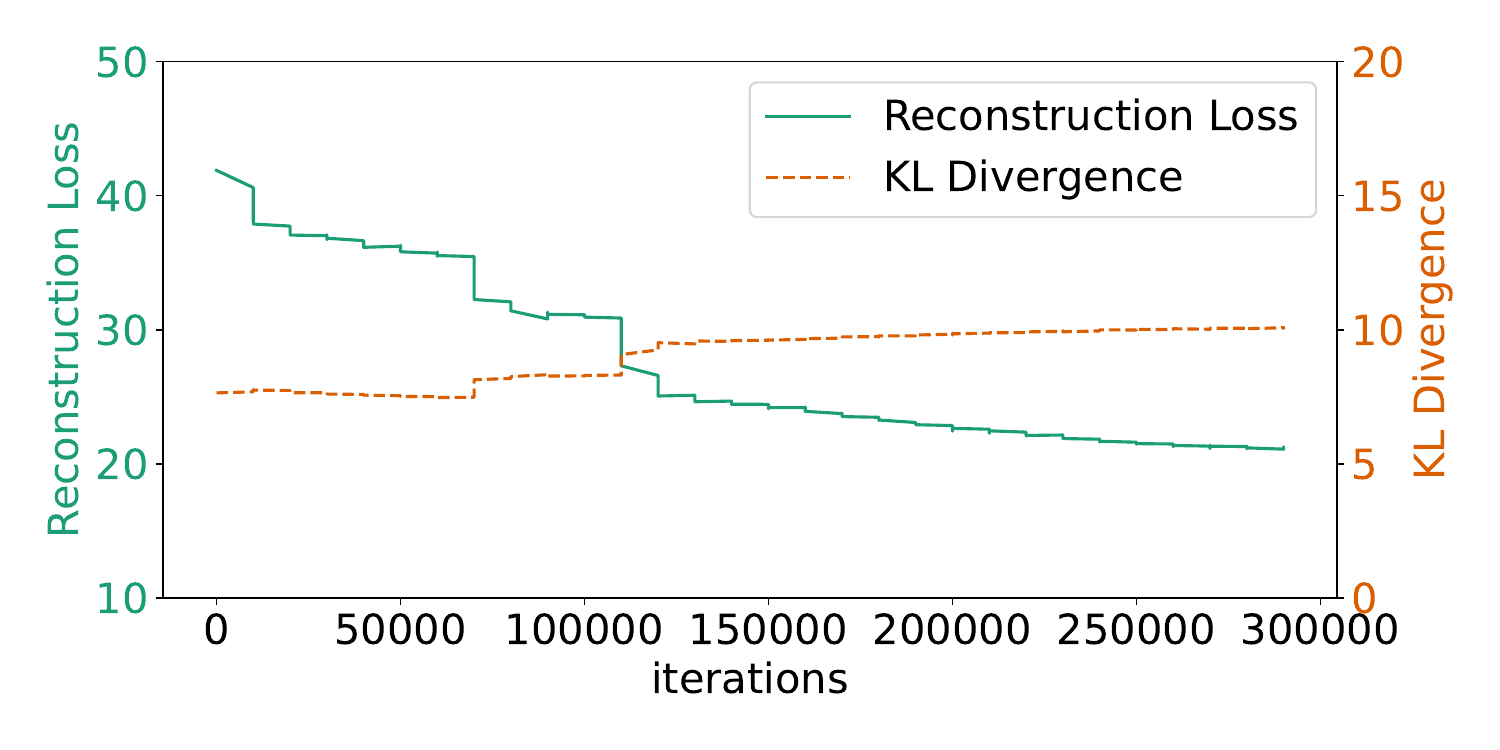}
         \caption{ }
         \label{fig:bvae-reconvskl}
    \end{subfigure}
    \caption{Training curves for reconstruction loss and KL-Divergence for (a) VAE and (b) $\beta$-VAE with $\beta=2$. (a) KL divergence increases during training due to the difference between the two loss terms. (b) $\beta$-VAE controls the difference between the two terms and the increase in the second term. Although the KL-divergence term does not increase as much as we observe in (a), it still does not decrease as expected. }
\end{figure*}

\section{A Critical Look at $\beta$ Variational Auto-Encoders ($\beta$-VAEs)} \label{sec:bvae}

$\beta$-VAEs  are one of the pioneering extensions of vanilla VAE \cite{higgins2016beta, burgess2018understanding}, that  generate a representation for the data. Employing an encoder-decoder architecture, $\beta$-VAEs estimate a Gaussian distribution of each source of variation, in the latent space.

Formally, given a dataset $\mathcal{D= \{X, W, Y\}}$, $\beta$-VAE obtains a representation (denoted by $\mathbf{z}_i$), which is sampled from learned Gaussian distributions, $\mathcal{N}(\mathbf{\mu}_i, \mathbf{v}_i)$, for $i= 1,..., L$, where $L$ is the number of hidden variables. 

An encoder-decoder architecture is used to estimate $\mathcal{N}(\mathbf{\mu}_i,\mathbf{v}_i)$, for the latent representation $\mathbf{z}_i$ and reconstructed data, $\mathbf{\Bar{x}}_i$. At training step, the encoder learns the parameters of the Gaussian distribution; $ \mathbf{\mu}_i$ and $ \mathbf{v}_i$. The latent representation $ \mathbf{z}_i$ of sample $ \mathbf{x}_i$ is then sampled from the learned distribution. 

$\beta$-VAE is trained to minimize the Evidence Lower BOund (ELBO):
\begin{eqnarray}\label{eq:bvae}\small
        \mathcal{L}_{\beta\textrm{-VAE}} & = & - E_{q (\mathbf{z}|\mathbf{x})}\Bigl[\log \underbrace{p (\mathbf{x}|\mathbf{z})}_{Decoder} \Bigr]  \\
                & & + \beta\cdot D_{KL}\Bigl(\underbrace{q(\mathbf{z}|\mathbf{x})}_{Encoder}\ \lVert\ p(\mathbf{z})\Bigr),
\end{eqnarray}
where the first term enhances the reconstruction quality of the images, compelling the generated images to match the original image, whereas the second term enforces the learned distributions to approach a predefined distribution, which is generally selected to be a Gaussian distribution with zero mean and unit variance ($p(\mathbf{z}) = \mathcal{N}(0,1)$). The distributions of the encoder $q(\mathbf{z}|\mathbf{x})$ and the decoder $p(\mathbf{x}|\mathbf{z})$ are parameterized by neural networks.

Notice that, for $\beta = 1,$ the model reduces to the original VAE. In this case, the  terms of the loss function, namely, mean squared error and KL divergence, becomes incompatible in terms of their order of magnitudes. Additionally, they pose different learning patterns with regards to the number of epochs and the learning rates for convergence. In order to compensate for these incompatibilities the KL-divergence term is scaled by a hyperparameter $\beta > 1$. This adjustment only harmonizes the numerical ranges  between the terms of the loss function, ignoring  the implicit learning difficulties, particularly in terms of training speeds, associated with each term.

\begin{figure*}[!ht]
    \centering
    \begin{subfigure}{0.24\linewidth}  
        \includegraphics[width=\textwidth]{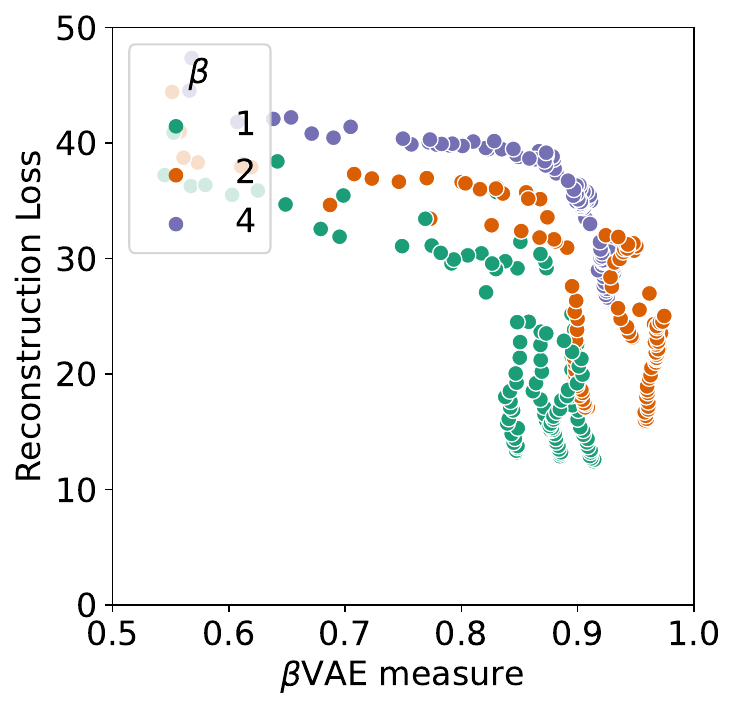}    
        \caption{ dSprites dataset }
          \label{fig:dsprites-beta-vs-others}
    \end{subfigure}
    \begin{subfigure}{0.24\linewidth}
        \includegraphics[width=\textwidth]{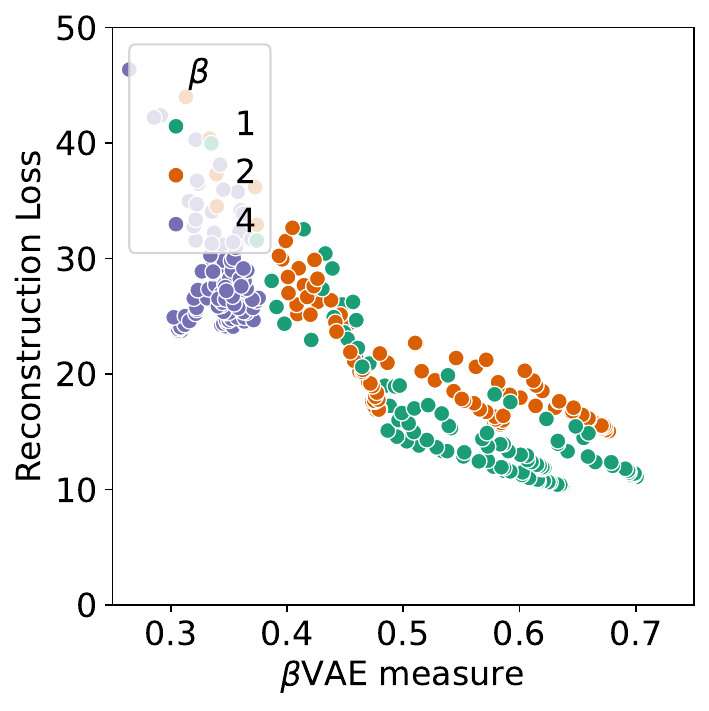}
         \caption{MPI3D dataset }
         \label{fig:mpi3d-beta-vs-others}
    \end{subfigure}
    \begin{subfigure}{0.24\linewidth}
        \includegraphics[width=\textwidth]{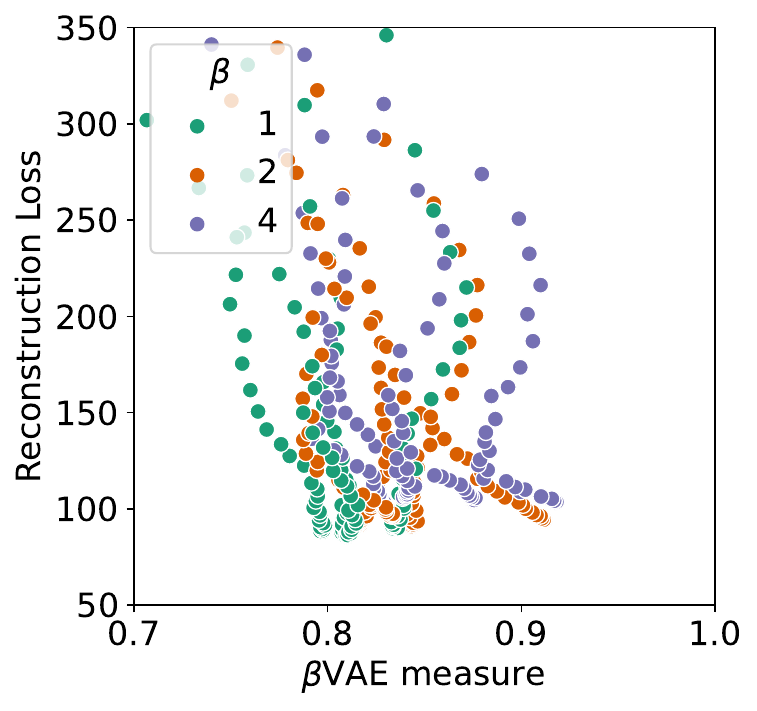}
         \caption{Falcor3D dataset }
         \label{fig:falcor-beta-vs-others}
    \end{subfigure}
    \begin{subfigure}{0.24\linewidth}
        \includegraphics[width=\textwidth]{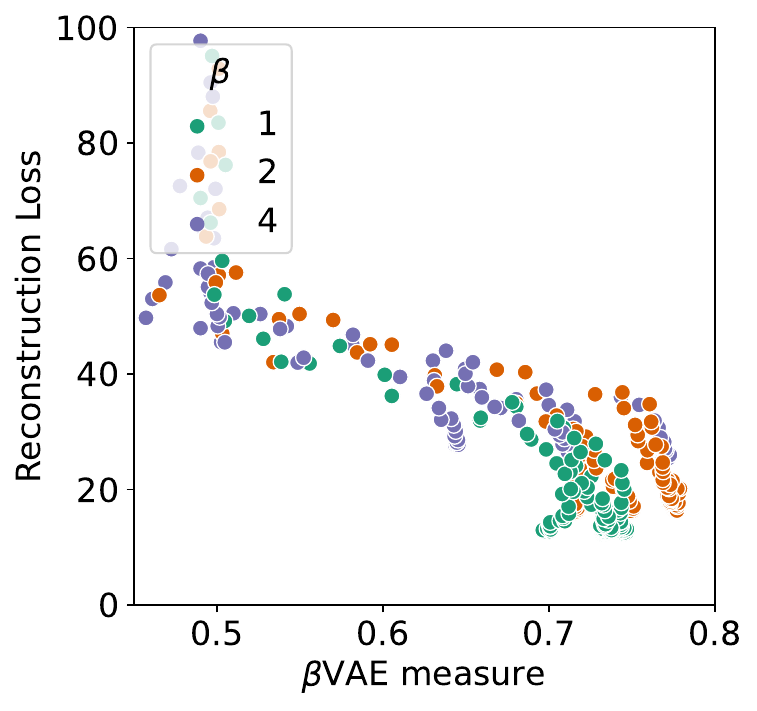}
         \caption{Isaac3D dataset }
         \label{fig:isaac-beta-vs-others}
    \end{subfigure}
    \caption{Reconstruction loss versus $\beta$-VAE disentanglement metric across different $\beta$ hyperparameters for (a) dSprites, (b) MPI3D, (c) Falcor3D, and (d) Isaac3D datasets. Each point represents a model trained with a different hyperparameter setup (altered learning rate, number of iterations, and batch size). Points are color-coded based on the value of $\beta$. }
    \label{fig:effetc-of-beta}
\end{figure*}

\subsection{Analysis on the Effect of the Hyperparameter $\beta$ on the Learning Dynamics.} \label{sec:analysis_on_beta} 

We start our research with an analysis on the effect of  the hyper parameter $\beta$ on the learning dynamics of $\beta$-VAE and provide a basis for our study. First, we investigate the learning dynamics between reconstruction and KL-divergence terms of Equation \ref{eq:bvae}. We analyze the relative changes in these terms during the training phase, for different values of the   hyperparameter $\beta$. Our observations are summarized below.

\noindent\textbf{Observation 1: The hyperparameter $\beta$ helps balancing the ranges of the reconstruction loss and the KL divergence.} 
Figure \ref{fig:vae-reconvskl} compares the reconstruction loss and KL divergence values during the training of a VAE, i.e., $\beta$-VAE with $\beta =1$. We observe that the values of the reconstruction loss  are relatively higher than the KL divergence values. As a consequence, the reconstruction loss dominates the loss function, and the contribution of the KL divergence term to the overall loss remains relatively small. Since the reconstruction loss dominates the overall loss, minimizing the KL divergence in parallel to the reconstruction loss becomes a challenge. This imbalance between the values of reconstruction loss and KL divergence term results in an increase in the  KL term, as the overall loss function converges to an optimal value. We believe that the leading cause of VAE's failure to disentangle the learned representation is the discrepancy between the values of the reconstruction loss and KL divergence term. Figure \ref{fig:bvae-reconvskl}, on the other hand, shows the behaviour of the same losses during the training phase for $\beta=2$ . The figure suggests that increasing the hyperparameter $\beta$  compensates for the discrepancy between the values of the two terms and dampens the increase of the KL term. 

\noindent\textbf{Observation 2: Increasing the hyper parameter $\beta$ lowers the reconstruction quality.} Figure \ref{fig:bvae-reconvskl} shows that weighing the importance of the disentanglement term by $\beta\ge 1$ also results in an increase in the reconstruction loss. This empirical finding suggests that, in most cases, the amount of disentanglement is inversely proportional to the amount of reconstruction loss.

We further  investigate the effect of $\beta$ on the relationship between disentanglement and reconstruction. We compare disentanglement properties on four datasets; dSprites \cite{dsprites17}, MPI3D \cite{gondal2019transfermpi3D}, Falcor3D \cite{isaac-falcor} and Isaac3D \cite{isaac-falcor} (details in Section \ref{sec:datasets}). We analyze the results of several runs covering a large hyperparameter space consisting of varying learning rates, batch sizes, and iteration counts (details in Section \ref{sec:model_details}). In this experiment, we have thoroughly investigated the relationship between the reconstruction and disentanglement terms by systematically varying the value of the hyperparameter $\beta$ across the  hyperparameters of the network. Figure \ref{fig:dsprites-beta-vs-others} shows the results of this analysis. The $x$ and $y$-axes show the {$\beta$-VAE disentanglement measure, and reconstruction loss, }respectively. Each point represents a model, trained with a different hyperparameter setup, with  learning rate, batch size, and number of iterations (see Section \ref{sec:model_details} for details). Points are color-coded based on the value of $\beta$. In order to visually observe  the effect of different $\beta$ values on the reconstruction loss and KL-divergence term, the values of other parameters are not specified in this Figure.

\noindent\textbf{Observation 3: Disentanglement also depends on hyperparameters other than $\beta$.} Figure \ref{fig:dsprites-beta-vs-others} shows that for the dSprites dataset, disentanglement vs. reconstruction dynamic highly depends on the selection of the hyperparameters, such as the learning rate or batch size. As we increase the  $\beta$ parameter, selecting the rest of the hyperparameters requires a larger search space. Thus, finding the optimal hyperparameter set becomes more difficult. We also observe that increasing the $\beta$ parameter from 1 to 2 gives better disentanglement values with the cost of higher reconstruction loss. These observations support the arguments of  \cite{chen2018isolating}, who claimed $\beta$-VAE increases the reconstruction loss. We observe similar behavior for the Falcor3D dataset (See Figure \ref{fig:falcor-beta-vs-others}).

\noindent\textbf{Observation 4: $\beta < 1$ can surprisingly increase disentanglement.} 
Further analysis on the MPI3D dataset reveals more insight into the effect of  $\beta$. We have conducted the same experiment on this dataset (see Figure \ref{fig:mpi3d-beta-vs-others}). The results on this dataset complement the dSprites experiments. Higgins et al. \cite{higgins2016beta} argued increasing the importance of KL Divergence yields better disentanglement and hence selected $\beta>1$. However, our results show that lower values of $\beta$ tend to provide better results in terms of lower reconstruction loss and higher disentanglement scores in this specific dataset. The flow of Figure \ref{fig:mpi3d-beta-vs-others} reveals that $\beta<1$ values should also be investigated. These findings support the results in \cite{fil2021beta}.  The behavior of the Isaac3D dataset is similar to the MPI3D dataset (See Figure \ref{fig:isaac-beta-vs-others}. We observe a decrease in the disentanglement measure as we increase the $\beta$ parameter.

\subsection{A Critique on $\beta$-VAE} 

$\beta$-VAE has three major drawbacks: First, it introduces a new parameter $\beta$ to hyperparameter space, which is to be tuned together with other hyperparameters of the model using expensive empirical methods in a large search space. Second, selecting a hyperparameter $\beta > 1$ to improve the amount of disentanglement may result in a relatively poor representation \cite{chen2018isolating} (Observation 2). Third, the ratio between reconstruction loss and the amount of disentanglement heavily depends on the network hyperparameters other than $\beta$ (Observation 3). The optimal value of $\beta$ can also be less than 1 for some datasets, further increasing the very large search space (Observation 4). 

The above drawbacks  highlight the importance of an effective learning model for the adjustment of the  loss function weights.  We believe that an optimal balance between the reconstruction loss and the degree of disentanglement requires a dynamic learning  model.  Our model L-VAE has a  self-learning mechanism that can  simultaneously optimize the model parameters and the  weights of the loss function without any restriction on the range of these weights,  as will be described in the next section.

\section{Learnable VAE (L-VAE)} \label{sec:lvae}
In this section, we describe the proposed  Learnable VAE (L-VAE), which circumvents some of  the drawbacks associated with the hyperparameter selection problem of $\beta$-VAE.  The proposed L-VAE can achieve lower reconstruction loss values than $\beta$-VAE, while producing better disentanglement scores. 

Our method is based on the multi-task learning method of Kendall et al. \cite{kendall2018multi}, where they learn the relative weights of different tasks in a loss term  by augmenting them to the optimizer. Without loss of generality, let us assume that  the overall loss function consists of two terms, $\mathcal{L}_0(\mathbf{x})$ and $\mathcal{L}_1(\mathbf{x})$, each of which are the functions of the input vector, $\mathbf x$. Then, the trade off between the terms of the loss function can be dynamically learned from the following analytical form:
\begin{equation}
    \mathcal{L}(\mathbf{x}) = \frac{1}{\sigma_0^2} \mathcal{L}_0(\mathbf{x}) + \frac{1}{\sigma_1^2} \mathcal{L}_1(\mathbf{x}) + \log (\sigma_0) + \log (\sigma_1),
    \label{eq:kendall}
\end{equation}
where $1 / \sigma_i^2$, for $i=0, 1$, are the balancing weights. The parameters $\sigma_i$ are optimized and updated with other learnable weights of the neural network. The last two terms of Equation \ref{eq:kendall} regularize the learned weights, $\sigma_0$ and $\sigma_1$. 

Building upon the same framework of Kendall et al. \cite{kendall2018multi}, we develop L-VAE such that it learns the relative weights of the terms of the VAE loss function by updating Equation \ref{eq:bvae}, as follows:
\begin{eqnarray}
    \mathcal{L}_{\textrm{L-VAE}} &=& - \frac{1}{\sigma_0^2}E_{q (\mathbf{z}|\mathbf{x})}\Bigl[\log p (\mathbf{x}|\mathbf{z}) \Bigr] \\
        & & + \frac{1}{\sigma_1^2}\mathcal{D}_{KL}\Bigl(q (\mathbf{z}|\mathbf{x})\ \lVert\ \mathbf{z}\Bigr) \nonumber 
          + \sum_{i=0,1} \sigma_i^2.
        \label{eq:lvae-loss}
\end{eqnarray}
In order to regulate the upper limit of the weight parameter $\sigma_i$ in Equation \ref{eq:lvae-loss}, we introduce a regularization term $\sum \sigma_i^2$. The $\sigma_i$ terms are added to  the parameter set of the optimizer and are estimated through simultaneous learning with the network parameters (refer to Section \ref{sec:sigma-curves} for the learning curves of $\sigma_i$).

Except the relative weights in the loss function, the rest of the network architecture is constructed exactly the same as  $\beta$-VAE. Therefore, a straightforward and fair comparison between $\beta$-VAE and L-VAE is possible through a simple conversion of the $\sigma_i$ parameters to the $\beta$ parameter. In  Section \ref{sec:quantitative-experiments}, we show  the correspondence of $\beta$ and $\sigma_i$ parameters. During the experiments,    we observe that, after the training phase, the ratio of the learned parameters,  $\frac{\sigma_0^2}{\sigma_1^2}$ in L-VAE is aligned with the  empirically tuned value of $\beta$ parameter in $\beta$-VAE.

\section{Experimental Setup}\label{sec:experiments}
In this section, we  describe the datasets, the encoder-decoder architectures, and the performance measures employed in our experiments.

\subsection{Datasets}\label{sec:datasets}
We conduct our experiments on four popularly used disentanglement data sets: 
\begin{enumerate} [1.]
\item {dSprites dataset \cite{dsprites17}}, which  is a 2D-shapes dataset with 700K 64x64 images containing white 2D shape images (heart, square, ellipse) on black background. There are five factors of variation to disentangle: shape, scale, orientation, and X, and Y positions of the object. 

\item{MPI3D-complex dataset \cite{gondal2019transfermpi3D}}, which consists of four real-world objects moving on a robotic arm leading to 460K 64$\times$64 colored images. There are seven factors of variation to disentangle: color shape, size, camera height, background color, and horizontal and vertical axes. We will refer to this dataset as MPI3D.

\item{Falcor3D dataset \cite{isaac-falcor}}, which  contains the images of a living room containing with different lighting conditions containing 233K 64$\times$64 images. There are seven factors of variation to disentangle: lighting intensity, directions x, y, and z of lighting, and x, y, and z coordinates of the camera position. 

\item{Isaac3D dataset \cite{isaac-falcor}}, which contains a robotic arm holding an object in a kitchen. Lighting conditions, camera position, and the position of the arm are altered. The dataset contains 737K 64$\times$64images. There are nine factors of variation to disentangle: Objects shape, scale, and color, wall color, camera height, robotic arms' x and y positions, lighting intensity and direction.

\end{enumerate}

We randomly split all datasets into training, validation, and test sets to conduct our experiments. The training set covers around 85\% of the dataset and the rest is equally split into test and validation sets.

\subsection{The Compared Methods}
We compare the proposed L-VAE with five well-known disentanglement  methods: 
\begin{enumerate} [1.]

\item Variational Autoencoder (VAE) \cite{kingma2013auto}, which optimizes reconstruction loss and KL divergence with equal weights.

\item $\beta$ Variational Autoencoder ($\beta$-VAE) \cite{higgins2016beta}, which utilizes an empirically tuned   hyperparameter $\beta$ to weight KL divergence.

\item ControlVAE \cite{shao2020controlvae,shao2021controlvae}, which learns the $\beta$ parameter, based on a PID control algorithm at each iteration.

\item DynamicVAE \cite{shao2022rethinking}, which is a slightly modified version of ControlVAE.

\item$\sigma$-VAE  \cite{rybkin2021simple}, which can be considered as an instantiation of our model L-VAE. It learns the weight of the reconstruction loss, whereas our model L-VAE learns the weights of both reconstruction loss and KL divergence.
\end{enumerate}

\subsection{The Details of L-VAE Model Architecture(s) }\label{sec:model_details}

We perform our experiments with an MLP Encoder-Decoder architecture for the dSprites dataset and a CNN Encoder-Decoder architecture for the MPI3D, Falcor3D, and Isaac3D datasets (see Table \ref{tab:architecture} for the architecture details). All of the architecture uses the ReLU activation function on all hidden layers and the Sigmoid activation function at the decoder output. We use the Adam optimizer \cite{kingma2014adam} with $\beta_1, \beta_2 = (0.9, 0.999)$, and $\epsilon=1e-08$. Mean Square Error (MSE) is used as the reconstruction loss.

\begin{table*}[h]
    \caption{The encoder and decoder architecture we used in our experiments. For fully connected (FC) layers, the number of output features is given in parentheses. For convolutional layers (Conv) and transposed convolutional layers (ConvT), the number of input and output features is also given in parentheses.}\label{tab:architecture}\small\centering
    \begin{tabular}{cc}
    \toprule
        Model & Model Details \\ \hline\hline 
         MLP $E(\cdot)$ & FC(1200), FC(1200), FC(2xLatent\_dim)\\ \cmidrule{2-2}
         MLP $D(\cdot)$ & FC(1200), FC(1200), FC(HxWxC)\\ \midrule
         CNN $E(\cdot)$ & \makecell{ Conv(32x4x4, stride=2),  Conv(32x4x4, stride=2),  Conv(64x4x4, stride=2), \\ Conv(64x4x4, stride=2),  Conv(32x4x4, stride=1), FC(2xLatent\_dim)} \\ \cmidrule{2-2}
         CNN $D(\cdot)$ & \makecell{FC(256), ConvT(64x4x4, stride=2), ConvT(64x4x4, stride=2), ConvT(32x4x4, stride=2),\\ConvT(32x4x4, stride=2), ConvT(3x4x4, stride=2), ConvT(3x4x4, stride=2)}\\ \bottomrule
    \end{tabular}
\end{table*}

We empirically tune the hyperparameters (namely, batch size, learning rate, and iteration count) for both the compared methods and L-VAE. Batch size is selected from \{32, 64, 128, 256\}, and the number of iterations from $\{1, 2, \ldots,30\}\times 10^4$. We used OneCycleLR optimization for learning rate where we initialized learning rate with 1e-5 and increased it to 1e-4 for the first 150000 iterations then we decrease the learning rate to 1e-6 using Cosine annealing strategy. For ControlVAE, we set the desired KL value to 18, $K_p$ to 0.0.01, $K_i$ to -0.001 and initialized $\beta$ with 0 (following  \cite{shao2020controlvae}). For DynamicVAE, we set the desired KL value to 18, $K_p$ to 0.0.01, $K_i$ to -0.005, and initialized $\beta$ with 150 (following \cite{shao2022rethinking}). 

Configuring experiments with this set of hyperparameters, VAE and L-VAE experiments are carried out independently for 120 models. For $\beta$-VAE, $\beta$ is selected from \{2,4\} (higher $\beta$ results in higher reconstruction losses for the datasets we have experimented on -- see Section \ref{sec:analysis_on_beta} for details), which leads to 240 independent models trained for $\beta$-VAE. We select the best set of hyperparameters for all models based on validation scores on the disentanglement measure with the $\beta$-VAE measure \cite{higgins2016beta}. 

We set the latent dimension size to five for the dSprites dataset, seven for the MPI3D and Falcor3D datasets, and nine for Isaac3D dataset 
following the number of labeled attributes provided with the datasets \cite{dsprites17, gondal2019transfermpi3D, isaac-falcor}. All the $\sigma_i$ values are initialized with one (1) for the L-VAE experiments.

\begin{figure*}[!ht]
    \centering
    \begin{subfigure}{0.48\linewidth}  
        \includegraphics[width=\textwidth]{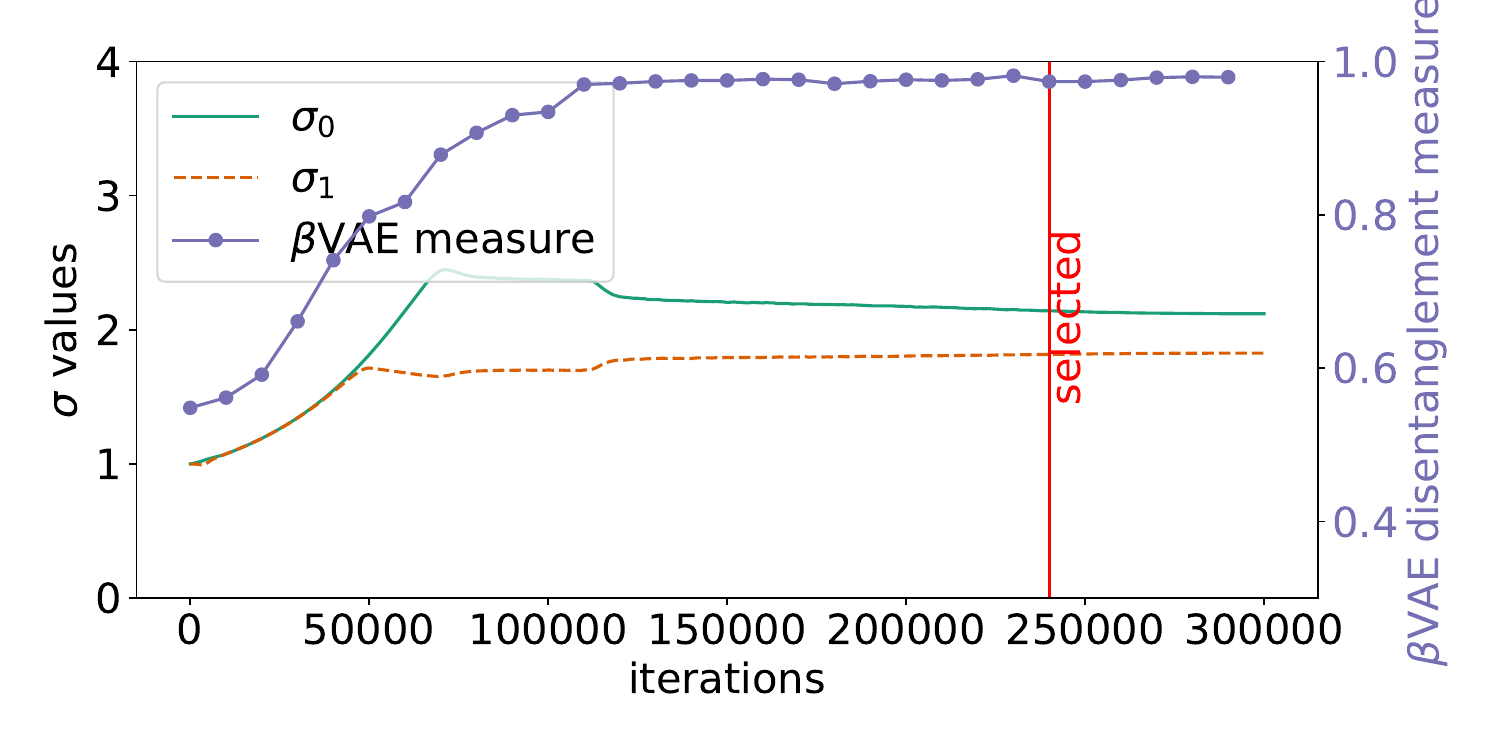}
         \caption{ dSprites dataset }
          \label{fig:dsprites-sigmas}
    \end{subfigure}
    \begin{subfigure}{0.48\linewidth}
        \includegraphics[width=\textwidth]{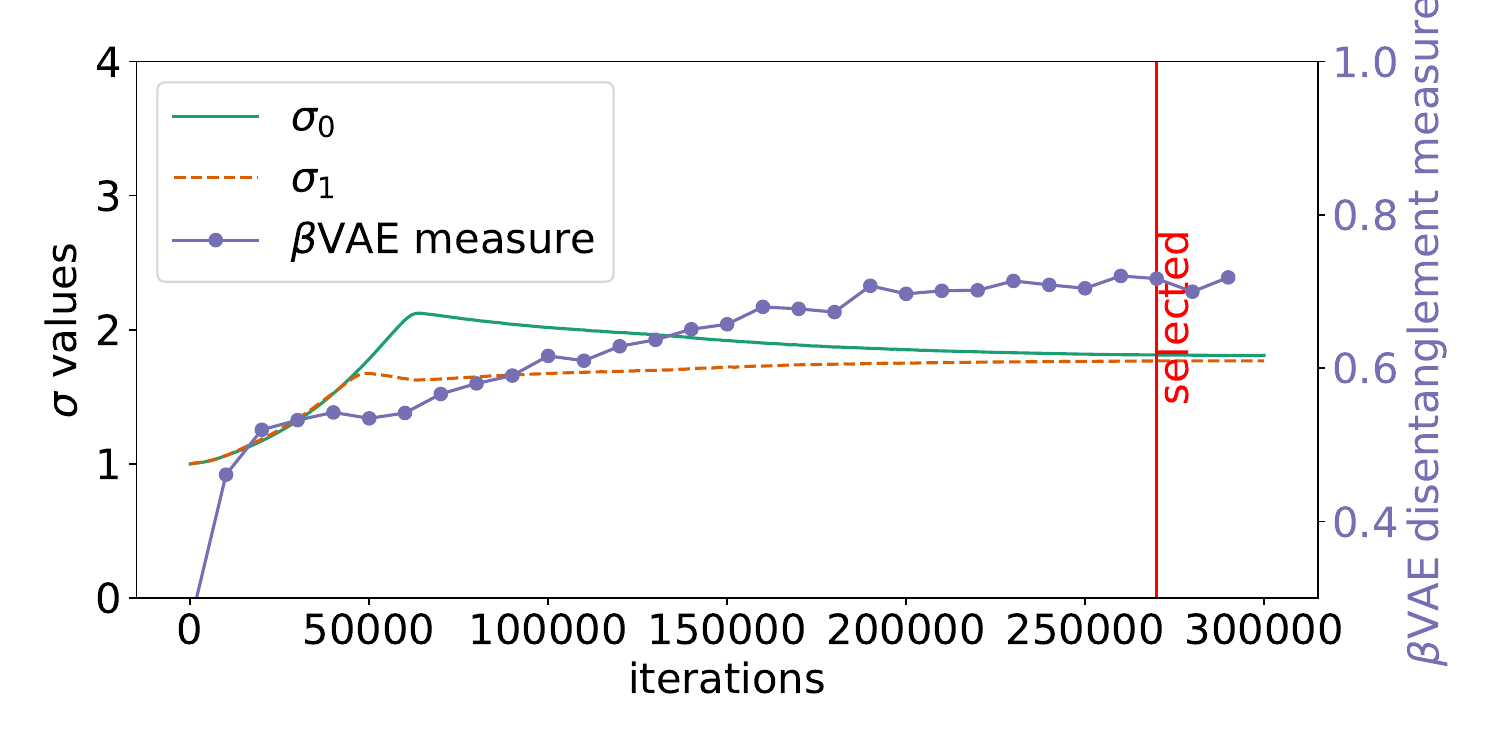}
         \caption{MPI3D dataset }
         \label{fig:mpi3d-sigmas}
    \end{subfigure}
    \begin{subfigure}{0.48\linewidth}
        \includegraphics[width=\textwidth]{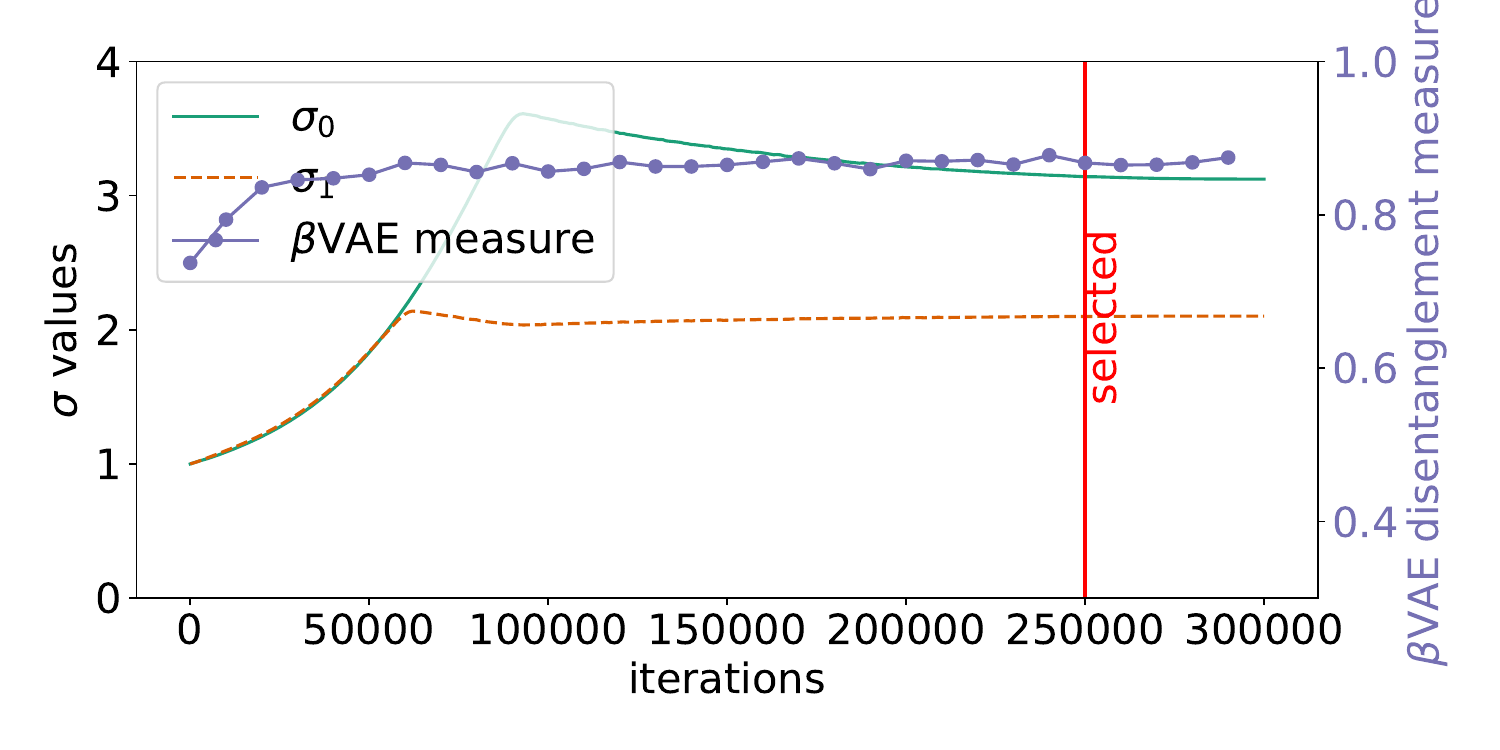}
         \caption{ Falcor3D dataset}
         \label{fig:falcor-sigmas}
    \end{subfigure}
    \begin{subfigure}{0.48\linewidth}
        \includegraphics[width=\textwidth]{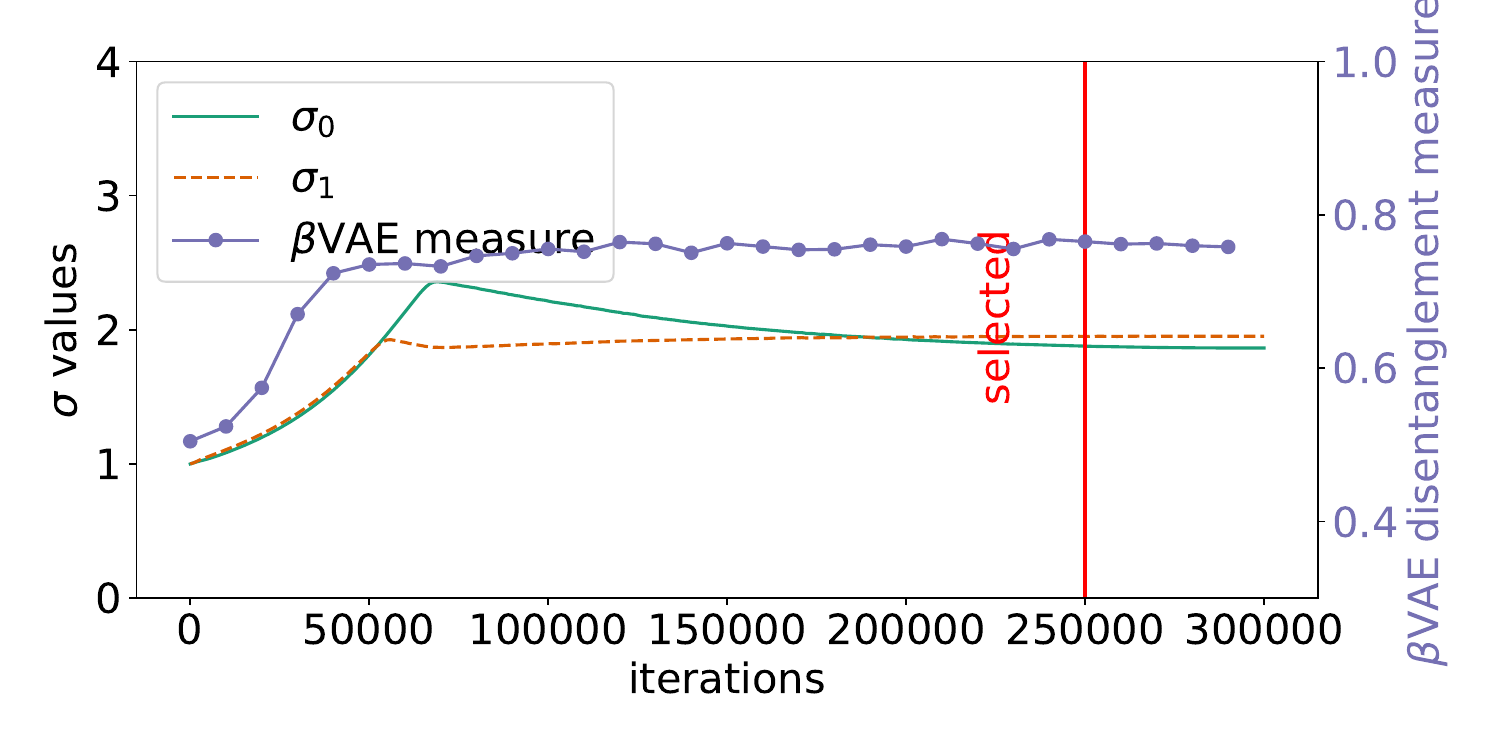}
         \caption{Isaac3D dataset }
         \label{fig:isaac-sigmas}
    \end{subfigure}
    \caption{ Training curves of the learned relative weights ($\sigma_0$ and $\sigma_1$) of LVAE model for (a) the dSprites, (b) the MPI3D (c) the Falcor3D, and (d) the Isaac3D datasets. The green line corresponds to the $\beta$-VAE disentanglement metric computed at every 10,000 iterations. We note that $\sigma_i$ reaches a peak around 50,000 iterations, after which the change decelerates. We determined the number of iterations (along with other hyper-parameters, learning rate, and batch size) based on the $\beta$-VAE disentanglement measure on validation dataset (see Section \ref{sec:selection-hyperparameters} for details). Red vertical lines show this selection. }
    \label{fig:sigma-learning-curves}
\end{figure*}

\subsection{Disentanglement Measures}\label{sec:measures}

In their review on disentanglement measures, Carbonneau et al. \cite{carbonneau2022measuring} argue that a disentangled model should be evaluated with respect to three major properties: Explicitness, compactness, and modularity. For evaluating the performances in our experiments, we selected the following disentanglement measures so that the three major properties are covered:

\begin{itemize}
    \item \textbf{Explicitness} evaluates a model's ability to recover the input from the representation. We have selected the \textit{explicitness score} \cite{ridgeway2018learning} to measure explicitness.
    
    \item \textbf{Compactness} signifies that a small portion of the representation (ideally one dimension) represents a single attribute. We have selected the \textit{Separated Attribute Predictability (SAP) score} \cite{kumar2017variational}, and \textit{Mutual Information Gap (MIG) score} \cite{chen2018isolating} to measure compactness.
    
    \item \textbf{Modularity} pertains to whether or not a change of a factor affects only a single dimension of the representation. We have selected the \textit{$\beta$-VAE} \cite{higgins2016beta} and the \textit{Factor VAE} \cite{kim2018disentangling} measures to quantify modularity.

\end{itemize} 
Some disentanglement measures encompass two or more of the above mentioned properties; in such instances, the measure is referred to as holistic  \cite{carbonneau2022measuring}. We have selected \textit{Interventional Robustness Score (IRS)} \cite{suter2019robustly}, which measures modularity and explicitness. 

Overall, we have selected six disentanglement measures for which we use the implementation provided by Carbonneau et al. \cite{carbonneau2022measuring}. Although these measures are commonly used in the literature, they define disentanglement differently or quantify different aspects of disentanglement. Thus, the ranking of each method may change across the datasets.

\subsection{Model Selection} \label{sec:selection-hyperparameters}
 As stated in Section \ref{sec:model_details}, we have conducted our experiments to cover a relatively large hyperparameter set (i.e., selection of batch size, learning rate, and iteration count). 
 There are two significant problems in determining the best set of hyperparameters of the models in the validation set: Firstly, finding a balance between the reconstruction loss and KL divergence in the overall loss function (Equation \ref{eq:lvae-loss}) is still an open research issue. Secondly, choosing a suitable measure for quantifying the amount of disentanglement brings a strong bias toward the selected measure. Therefore, empirically finding an "optimal"  hyperparameter set is not a well-defined problem in disentangled representation learning. For the sake of being fair in comparing the proposed L-VAE  with the other VAE models, we selected the hyperparameters based on the highest $\beta$-VAE score, which is proposed along with $\beta$-VAE in \cite{higgins2016beta}. We use the best scores achieved on the validation set for the hyperparameter selection for all methods.

\begin{table*}[!ht]
    \centering
     \small 
     \caption{Quantitative comparison of VAE, $\beta$-VAE, ControlVAE, DynamicVAE, $\sigma$-VAE, and L-VAE based on reconstruction loss and six disentanglement measures across four datasets. }
    \label{tab:results}
    \begin{tabular}{cllccccccc}
    \toprule
         \rotatebox{90}{Dataset}& \rotatebox{90}{Model} & \rotatebox{90}{Parameters} & \rotatebox{90}{Reconstruction} $\downarrow$ & \rotatebox{90}{$\beta$-VAE} $\uparrow$ & \rotatebox{90}{FactorVAE} $\uparrow$ & \rotatebox{90}{Explicitness} $\uparrow$ & \rotatebox{90}{IRS} $\uparrow$  & \rotatebox{90}{MIG} $\uparrow$&\rotatebox{90}{SAP} $\uparrow$\\ \midrule\midrule
\multirow{6}{*}{\STAB{\rotatebox{90}{dSprites}}}&  VAE \cite{kingma2014adam}&   $\beta=1$&\textbf{12.46}&0.91&0.62&0.50&0.34&0.11&0.09\\ \cmidrule{2-10}
         &  $\beta$-VAE \cite{higgins2016beta}&   $\beta=2$&25.04&\textbf{0.97}&\underline{0.74}&\underline{0.58}&\underline{0.59}&0.30&0.27\\ \cmidrule{2-10}
         &  ControlVAE \cite{shao2020controlvae}&  - &28.00&\underline{0.96}&0.71&0.57&0.57&\textbf{0.34}&\textbf{0.34}\\ \cmidrule{2-10}
         &  DynamicVAE \cite{shao2022rethinking}&  - &31.57&0.93&0.71&0.55&0.53&\underline{0.31}&0.28\\ \cmidrule{2-10}
         &  $\sigma$-VAE \cite{rybkin2021simple}&  - &29.21&0.85&0.61&0.51&0.38&0.10&0.05\\ \cmidrule{2-10}
         &  L-VAE&   \makecell{$\hat{\beta} = \frac{\sigma_0^2}{\sigma_1^2}=1.39$}&\underline{21.14}&\textbf{0.97}&\textbf{0.77}&\textbf{0.59}&\textbf{0.63}&0.30&\underline{0.29}\\ \midrule\midrule
\multirow{6}{*}{\STAB{\rotatebox[origin=c]{90}{MPI3D}}}&  VAE&   $\beta=1$&11.21&\underline{0.70}&0.32&\underline{0.41}&\underline{0.32}&\textbf{0.16}&\underline{0.18}\\ \cmidrule{2-10}
         &  $\beta$-VAE&   $\beta=2$&15.31&0.67&\underline{0.44}&0.36&0.31&\textbf{0.16}&0.17\\ \cmidrule{2-10}
         & ControlVAE& - &14.34&0.67&0.42&0.35&0.32&\underline{0.13}&0.12\\ \cmidrule{2-10}
         & DynamicVAE&  -&15.81&0.54&0.35&0.30&0.30&0.12&0.13\\ \cmidrule{2-10}
         & $\sigma$-VAE&  -&\textbf{5.42}&0.66&0.38&\textbf{0.48}&\textbf{0.44}&0.03&0.02\\ \cmidrule{2-10}
         & L-VAE& \makecell{$\hat{\beta} = \frac{\sigma_0^2}{\sigma_1^2}=1.05$ }&\underline{10.79}&\textbf{0.71}&\textbf{0.46}&0.39&\underline{0.32}&\textbf{0.16}&\textbf{0.20}\\ \midrule\midrule
\multirow{6}{*}{\rotatebox{90}{Falcor3D}}&VAE&   $\beta=1$&215.03&0.87&0.42&0.58&\underline{0.32}&0.04&0.03\\ \cmidrule{2-10}
        &  $\beta$-VAE&   $\beta=4$  &105.17&\textbf{0.92}&\textbf{0.61}&\textbf{0.66}&\textbf{0.33}&\textbf{0.07}&0.05\\ \cmidrule{2-10}
        & ControlVAE& - &187.41&0.78&0.40&0.40&0.20&\underline{0.06}&\textbf{0.07}\\ \cmidrule{2-10}
        & DynamicVAE&  -&216.87&0.72&0.30&0.38&0.16&0.06&\underline{0.06}\\ \cmidrule{2-10}
        & $\sigma$-VAE&  -&\textbf{78.82}&\underline{0.89}&0.39&\underline{0.61}&0.30&0.04&0.03\\ \cmidrule{2-10}
        & L-VAE& \makecell{$\hat{\beta} = \frac{\sigma_0^2}{\sigma_1^2}=2.34$ }&\underline{97.97}&0.88&\underline{0.47}&\textbf{0.66}&0.30&0.05&0.05\\ \midrule\midrule
\multirow{6}{*}{\rotatebox{90}{Isaac3D}}& VAE& $\beta=1$ &\underline{13.37}&0.75&0.48&\underline{0.56}&0.24&0.07&0.07\\ \cmidrule{2-10}
       & $\beta$-VAE& $\beta=2$ &17.08&\textbf{0.78}&\underline{0.52}&0.54&0.31&\textbf{0.22}&\textbf{0.19}\\ \cmidrule{2-10}
       & ControlVAE& -&27.45&0.60&0.34&0.42&0.24&0.13&0.14\\ \cmidrule{2-10}
       & DynamicVAE& -&35.55&0.49&0.27&0.37&0.21&0.13&0.13\\ \cmidrule{2-10}
       & $\sigma$-VAE&  -&23.09&0.67&0.49&0.55&\textbf{0.33}&0.05&0.04\\ \cmidrule{2-10}
       & L-VAE& \makecell{$\hat{\beta} = \frac{\sigma_0^2}{\sigma_1^2} = 0.95$ }&\textbf{12.97}&\underline{0.76}&\textbf{0.61}&\textbf{0.57}&\underline{0.32}&\textbf{0.17}&\underline{0.15}\\ \bottomrule
    \end{tabular}
\end{table*}

\section{Quantitative Experiments} \label{sec:quantitative-experiments}

In this section, first, we analyse the learned weights of the proposed L-VAE. Then, we evaluate the performance of our L-VAE model in comparison with the baseline VAE models. We measure the performances by employing the reconstruction loss and disentanglement measures, which are selected from the literature, as explained in the previous section.  Finally, we carry out an ablation study to explore the effects of a dynamic learning strategy proposed in this paper.

\subsection{Experiment 1: Convergence of the  Weights Learned by L-VAE} \label{sec:sigma-curves}

In this set of experiments, we analyze how the parameters $\sigma_0$ and $\sigma_1$ (Equation \ref{eq:lvae-loss}) change over time during the learning phase. In all experiments,  we select the $\sigma_i$ values that maximize the $\beta$-VAE score.  

During the  derivations of L-VAE, we mentioned that there is a correspondence between the empirically tuned $\beta$ parameter of 
$\beta$-VAE model and the optimal ratio of the  learned parameter of L-VAE model,
$$\hat{\beta} = \frac{\sigma_0^2}{\sigma_1^2}.$$
 Hence, we shall also investigate the behavior of the optimal  ratio $\hat{\beta}$ for each dataset.

Figure \ref{fig:sigma-learning-curves} shows the learning curves of $\sigma_i$ parameters. We notice that, in the dSprites dataset, $\sigma_i$ exhibits a steep ascent until 50K iterations, reaching a peak value, after which the change attenuates. However, the $\beta$-VAE score slightly decreases as $\sigma_0$ decreases. During the cross-validation step, the  $\sigma_i$ values, obtained  at 50Kth iteration is selected. These values corresponds to the maximum $\beta$-VAE score.

The learning curves of $\sigma_i$ parameters for the rest of the datasets show similar behaviors. However, the $\beta$-VAE scores keep increasing during the training phase. As  $\beta$-VAE scores increase, we observe that  the ratio of $\sigma_0^2 /\sigma_1^2$  also changes: For the dSprites, and Falcor3D datasets,  $\sigma_0$ surpasses $\sigma_1$, yielding the generally practiced estimation for 

$$\hat{\beta} = \frac{\sigma_0}{\sigma_1}> 1.$$

  On the other hand,  in the MPI3D, and Isaac3D dataset, $\sigma_1$ overtakes $\sigma_0$, yielding an unusual estimation for

$$\hat{\beta} = \frac{\sigma_0}{\sigma_1} \leq 1.$$

The above result is rather counter-intuitive, considering the fact that disentanglement is accentuated for $\hat{\beta} > 1.$

\subsection{Experiment 2: Comparison with Baseline Methods}

In this set of experiments, we compare VAE, $\beta$-VAE, ControlVAE, DynamicVAE, $\sigma$-VAE, (see Section \ref{sec:auto-tuning} for a brief description of the methods) and L-VAE based on the MSE reconstruction loss and the six disentanglement measures \cite{carbonneau2022measuring} mentioned in the previous subsection. Table \ref{tab:results} shows the results for all four datasets. For $\beta$-VAE, the $\beta$ value is determined through a hyperparameter search process described in Section \ref{sec:selection-hyperparameters}, whereas, for L-VAE, the values of $\sigma_i$ represent the learned values. 

First, we compare the learned value of $\hat{\beta} = \sigma_0^2 /\sigma_1^2$ with the empirically tuned $\beta$ parameter. The learned value of $\hat{\beta}$ is 1.39, 1.05, 2.34, and 0.95 for dSprites, MPI3D, Falcor3D, and Isaac3D datasets, respectively. These values are consistent with our findings in Figure \ref{fig:effetc-of-beta}, which suggests that the optimal value of $\beta$ might be lower than 1 for the Isaac3D dataset. These results are aligned with the ratio of the weights  $\hat{\beta} = \sigma_0^2 /\sigma_1^2$ learned by L-VAE.

To assess the disentanglement performances of the different models, we examine their disentanglement properties (modularity, explicitness, and compactness) separately. 
Regarding modularity (looking at the $\beta$-VAE and FactorVAE measures), L-VAE achieves better scores in the dSprites and MPI3D datasets. However, it performs on par in other datasets. Similarly, with respect to the explicitness score, L-VAE achieves the best performance for the dSprites, Falcor3D, and Isaac3D datasets.  Regarding the compactness of representations (the SAP score), L-VAE achieves better compactness in the MPI3D dataset. Finally, we compare the models with the holistic IRS measure, which combines modularity and explicitness properties. The best score is achieved with the L-VAE model in dSprites dataset.

Overall, the results suggest that L-VAE can learn weights ($\sigma_i$) consistent with our preliminary observations where we show that increasing $\beta$ can lead to higher reconstruction loss and $\beta<1$ can produce higher disentanglement scores (See Observations 2 and 4 in Section \ref{sec:analysis_on_beta}, and Figure \ref{fig:effetc-of-beta}). Moreover, L-VAE generally produces better or on par reconstructions. 
Although the six disentanglement measures do not suggest a consistent ordering among the methods, we see that L-VAE achieves superior or on par performance with respect to many measures on all datasets.

\begin{figure*}[!t]
\centering
\fbox{
    \begin{subfigure}[t]{0.27\linewidth}  
        \includegraphics[width=\textwidth]{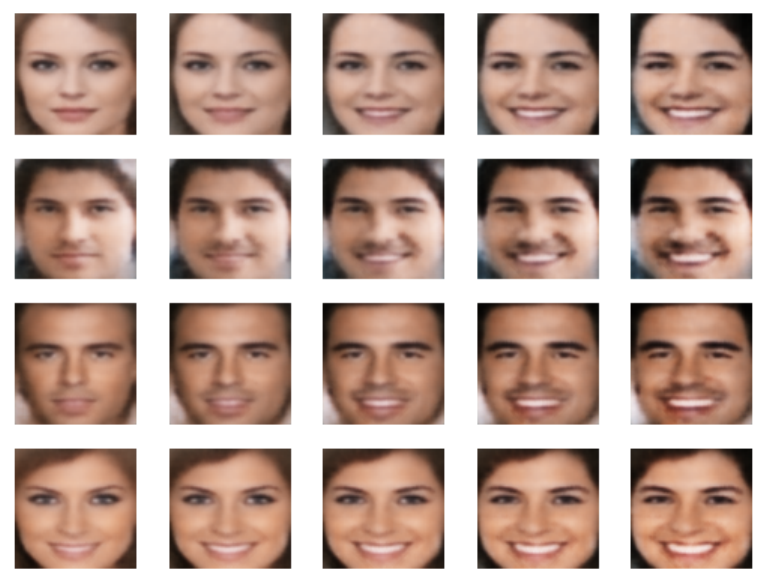} 
        \caption{Smile [0,4]}
        \label{fig:a}
    \end{subfigure}
    }
    \fbox{
    \begin{subfigure}[t]{0.27\linewidth}  
        \includegraphics[width=\textwidth]{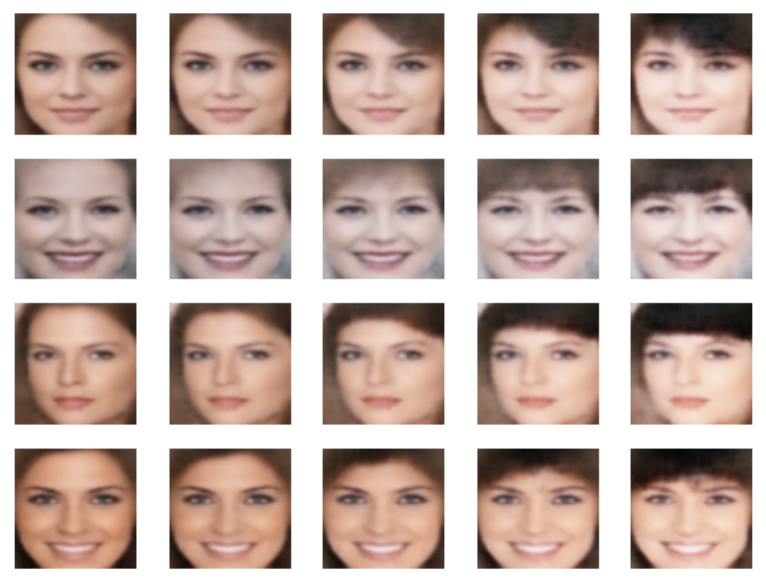}
        \caption{Bangs [0,4]}
        \label{fig:b}
    \end{subfigure}
    }
    \fbox{
    \begin{subfigure}[t]{0.27\linewidth}  
        \includegraphics[width=\textwidth]{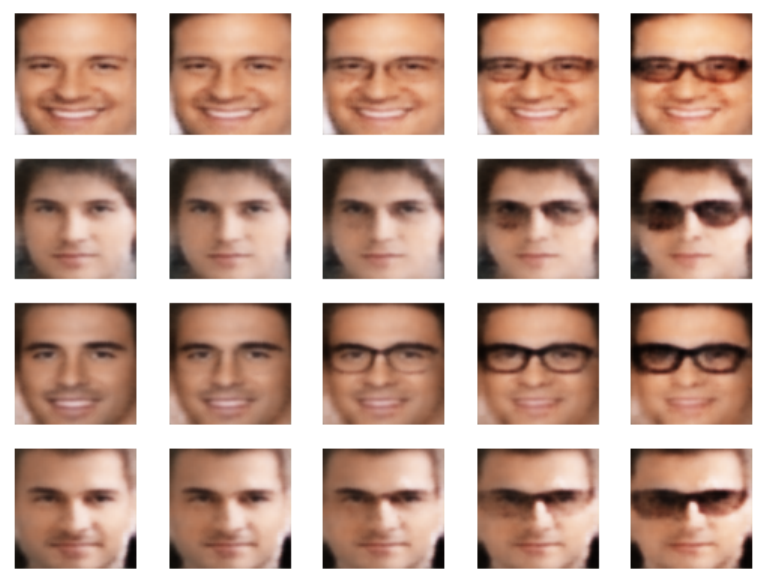}
        \caption{Eyeglasses [0,4]}
        \label{fig:c}
    \end{subfigure}
    }
    \fbox{
    \begin{subfigure}[t]{0.27\linewidth}  
        \includegraphics[width=\textwidth]{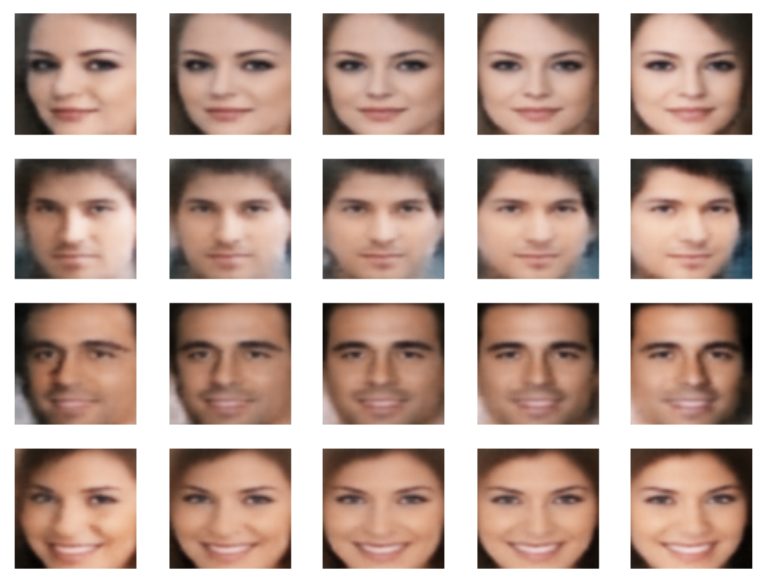}
        \caption{Camera Position [-2,2]}
        \label{fig:d}
    \end{subfigure}
    }
    \fbox{
    \begin{subfigure}[t]{0.27\linewidth}  
        \includegraphics[width=\textwidth]{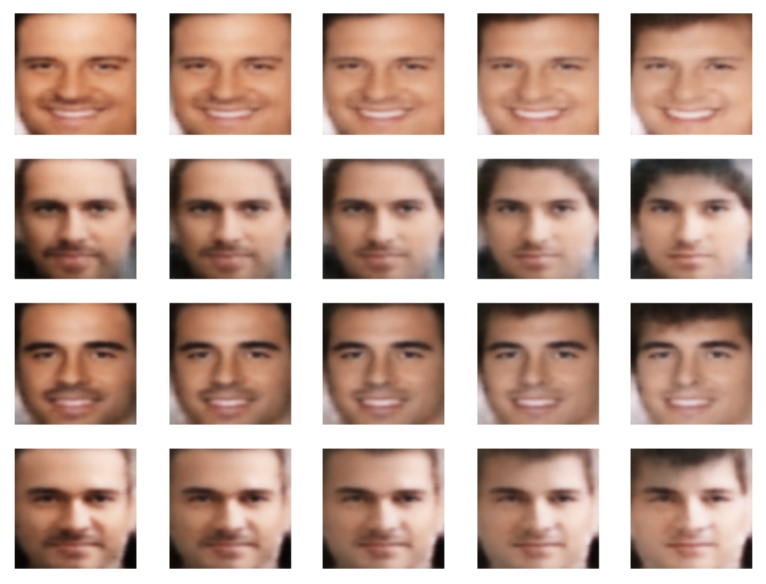}
        \caption{Receding Hairline [-2,2]}
        \label{fig:e}
    \end{subfigure}
    }
    \fbox{
     \begin{subfigure}[t]{0.27\linewidth}  
        \includegraphics[width=\textwidth]{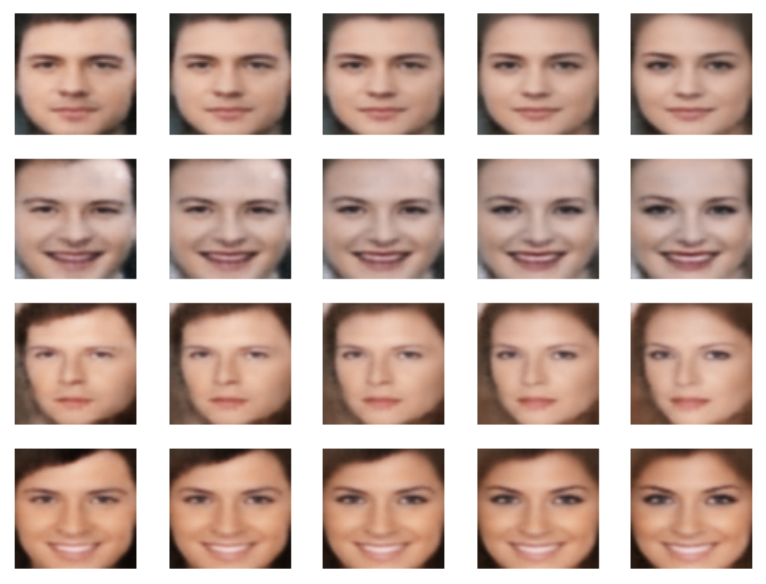}
        \caption{Gender [-4,0]}
        \label{fig:f}
    \end{subfigure}} 
    \fbox{
    \begin{subfigure}[t]{0.27\linewidth}  
        \includegraphics[width=\textwidth]{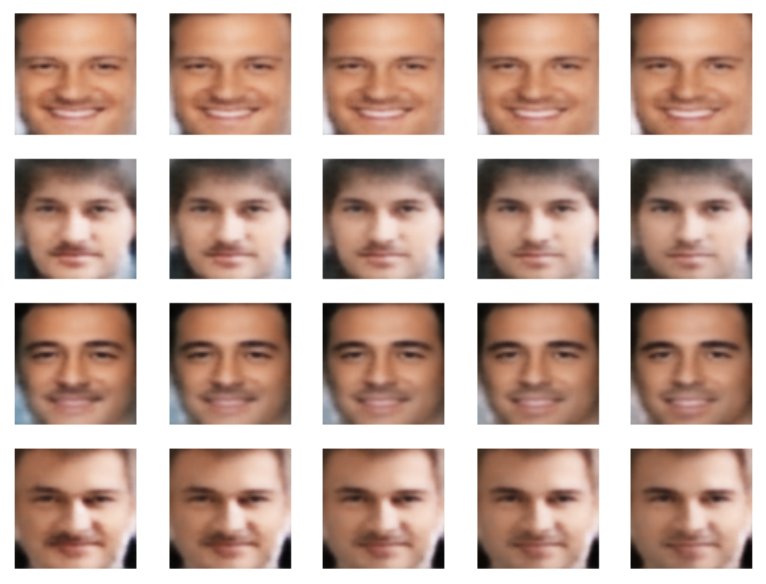}
        \caption{Mustache [-4,0]}
        \label{fig:g}
    \end{subfigure}
    }
    \fbox{
    \begin{subfigure}[t]{0.27\linewidth}  
        \includegraphics[width=\textwidth]{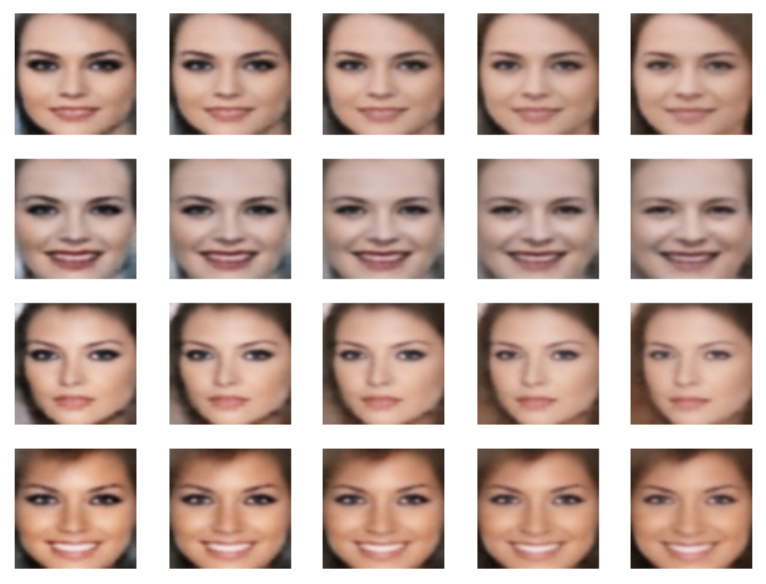}
        \caption{Make Up [-4,0]}
        \label{fig:h}
    \end{subfigure}
    }
    \fbox{
    \begin{subfigure}[t]{0.27\linewidth}  
        \includegraphics[width=\textwidth]{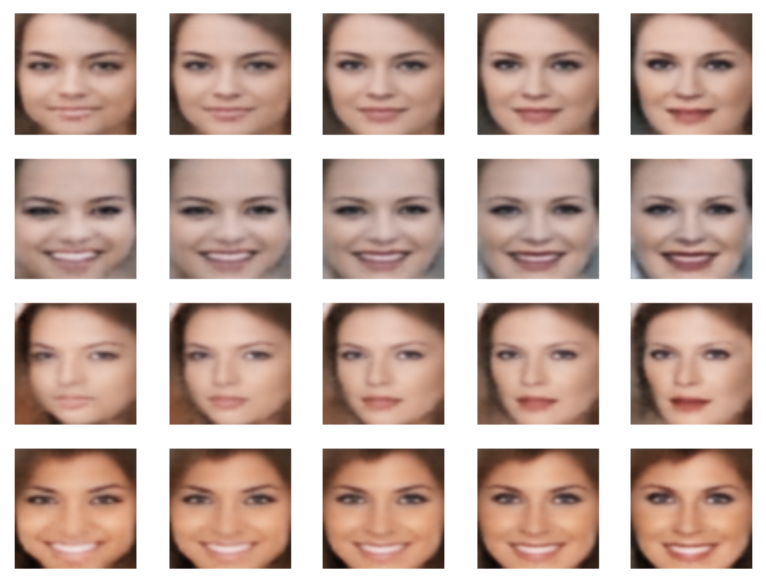}
        \caption{Age [-4,4]}
        \label{fig:i}
    \end{subfigure}
    }
    \caption{Latent traversals of the L-VAE model for the CelebA dataset. We select a single dimension and acquire reconstructions while altering its value. Note that other dimensions are kept unchanged in these experimens. Each subfigure corresponds to the traversal of a single latent dimension. The ranges of the alteration are shown in brackets.}
    \label{fig:latents}
\end{figure*}
\subsection{Experiment 3: Ablation Study}

In order to further investigate the effect of learning the hyperparameters of L-VAE on the reconstruction loss and the disentanglement term, we train a standalone $\beta$-VAE model with the learned weights as $\beta=\sigma_0^2 / \sigma_1^2$ as an ablation study. The results of these experiments are shown in Table \ref{tab:ablation} for all datasets.

Notice that the reconstruction loss of the of the $\beta$-VAE trained by the learned parameter $\hat\beta$ a is {close to} that of the  L-VAE, except for the dSprites dataset. Moreover, training $\beta$-VAE with the $\hat{\beta}$ weights learned with L-VAE exhibits similar or inferior performance than L-VAE in terms of the $\beta$-VAE measure in the dSprites dataset, which signifies the importance of a dynamic weighting strategy. 


\begin{table*}[ht]
\centering\small
\caption{An ablation study, where we train a standalone $\beta$-VAE with  $\beta=\hat{\beta} = \sigma_0^2 /\sigma_1^2$. The weights $\sigma_i$ are learned from the suggested L-VAE model. The value of $\hat{\beta} = \sigma_0^2 /\sigma_1^2$ are given in the second column. This setup results in a higher reconstruction loss and lower $\beta$-VAE measure than we have obtained with L-VAE, indicating the importance of a dynamic learning strategy. }
\label{tab:ablation}
\begin{tabular}{cccccccccc}
\toprule
\rotatebox{90}{Dataset}& \rotatebox{90}{Model} & \rotatebox{90}{Parameters} & \rotatebox{90}{Reconstruction} $\downarrow$ &  \rotatebox{90}{$\beta$-VAE} $\uparrow$ & \rotatebox{90}{FactorVAE} $\uparrow$ & \rotatebox{90}{Explicitness} $\uparrow$ & \rotatebox{90}{IRS} $\uparrow$  & \rotatebox{90}{MIG} $\uparrow$&\rotatebox{90}{SAP} $\uparrow$\\ \midrule\midrule
\multirow{2}{*}{\rotatebox{90}{dSprites}} &$\beta$VAE& $\beta = \hat{\beta} $&\textbf{18.13}&0.96&0.74&0.49&0.50&0.16&0.07\\ \cmidrule{2-10}
&  L-VAE&   \makecell{$\hat{\beta} = \frac{\sigma_0^2}{\sigma_1^2}=1.39$}&21.14&\textbf{0.97}&\textbf{0.77}&\textbf{0.59}&\textbf{0.63}&\textbf{0.30}&\textbf{0.29}\\  \midrule\midrule
\multirow{2}{*}{\rotatebox{90}{MPI3D}}  &$\beta$VAE& $\beta=\hat{\beta}$&\textbf{10.63}&\textbf{0.74}&\textbf{0.46}&\textbf{0.39}&\textbf{0.32}&\textbf{0.17}&\textbf{0.21}\\ \cmidrule{2-10}
& L-VAE& \makecell{$\hat{\beta} = \frac{\sigma_0^2}{\sigma_1^2}=1.05$ }&10.79&0.71&\textbf{0.46}&\textbf{0.39}&\textbf{0.32}&0.16&0.20\\ \midrule\midrule
 \multirow{2}{*}{\rotatebox{90}{Falcor3D}} &$\beta$VAE& $\beta=\hat{\beta}$&98.36&\textbf{0.93}&\textbf{0.65}&0.61&\textbf{0.35}&\textbf{0.12}&\textbf{0.09}\\ \cmidrule{2-10}
 & L-VAE& \makecell{$\hat{\beta} = \frac{\sigma_0^2}{\sigma_1^2}=2.34$ }&\textbf{97.97}&0.88&0.47&\textbf{0.66}&0.30&0.05&0.05\\  \midrule\midrule
\multirow{2}{*}{\rotatebox{90}{Isaac3D}} &$\beta$VAE& $\beta=\hat{\beta}$&\textbf{12.35}&\textbf{0.77}&0.52&\textbf{0.58}&0.25&0.07&0.06\\ \cmidrule{2-10}
& L-VAE& \makecell{$\hat{\beta} = \frac{\sigma_0^2}{\sigma_1^2} = 0.95$ }&12.97&0.76&\textbf{0.61}&0.57&\textbf{0.32}&\textbf{0.17}&\textbf{0.15}\\ \bottomrule
\end{tabular}
\end{table*}

\section{Qualitative Experiments}

Following the common practice in the literature \cite{shao2022rethinking, chen2018isolating, higgins2016beta}, we provide qualitative results for L-VAE on the CelebA dataset \cite{celeba}. CelebA consists of 202K images of celebrity faces labeled with facial attributes. There are 40 facial attribute labels, such as baldness, wearing eyeglasses, and pale skin. We cropped the background from the images and downscaled the dataset to size 128$\times$128 following \cite{chen2018isolating}. Furthermore, we have used 12\% of the dataset to train the L-VAE model for simplicity. We have trained the CNN Encoder-Decoder architecture in Table \ref{tab:architecture}, with the batch size of 128 and the learning rate of 1e-4 for 1M iterations. 

In order to assess the disentanglement ability of L-VAE qualitatively, we analyze the latent-space traversals through reconstructions as suggested in the literature \cite{shao2022rethinking, chen2018isolating, higgins2016beta}. After encoding sample images, we acquire latent representations for images. We select a specific latent dimension and alter its value while keeping other dimensions unchanged. 

Figure \ref{fig:latents} shows the results of latent traversals for nine different dimensions. As shown in the Figure, altering a single latent dimension consequently changes a single attribute in facial images; for example, Figure \ref{fig:i} shows that people are aged as we alter the value of a specific dimension.

\section{Conclusion}

In this study, we propose an extension to the classical  $\beta$-VAE. The proposed model, called, L-VAE,  dynamically learns the relative weights of reconstruction loss and KL divergence term. Our study is inspired from  the findings of Locatello et al. \cite{locatello2019challenging}, which  argue that hyperparameter selection has more impact on the disentanglement properties compared to the  model selection itself. Hence, we suggest a straightforward and efficient algorithm for  learning the hyperparameters of the loss function. The proposed optimization methodologies are also applicable to similar deep learning models.

The foundation of our study rests upon the power of $\beta$-VAE in learning a disentangled representation with two major challenges:

First of all, $\beta$-VAE  increases the disentanglement abilities of the VAE model, at the cost of  decreased reconstruction quality. Secondly, the introduction of the hyperparameter $\beta$ expands the search space of hyperparameters. The proposed L-VAE provides partial remedies to these problems.

 L-VAE can estimate  the optimal ratio of weights concerning the trade-off between reconstruction loss and the disentanglement of the learned representation  without introducing additional hyperparameters to the model. Our experiments  demonstrate that the L-VAE  outperforms or match the state of the art methods in disentanglement. Moreover,  L-VAE  learns    remarkable disentangled representations, while  yielding substantially low reconstruction losses.

L-VAE can learn the weights of the losses without any assumptions on the dynamic range of the hyperparameters. A common assumption made in the literature is to select a higher weight for the KL divergence to ensure better disentanglement\cite{higgins2016beta}. However, based on our results, we showed that selecting a smaller weight for KL divergence may results in higher disentanglement scores in some datasets, such as  Isaac3D dataset. L-VAE learns the optimal weights, which establish a very sensitive balance between the reconstruction loss and KL divergence term without modifications to the model or the hyperparameter space.

We made an interesting comparison  between the  $\beta$ VAE  and our  L-VAE:   In our experimentation, we  train the classical $\beta$-VAE model with the weights learned at the output of the proposed  L-VAE model. $\beta$-VAE model achieves better disentanglement scores with the learned $\beta$ values, with the cost of an increased  reconstruction loss. This observation demonstrates the significance of  the dynamic learning process suggested in L-VAE.

\section{Acknowledgments}

We also gratefully acknowledge the computational resources kindly provided METU ImageLab and METU Robotics and Artificial Intelligence Technologies Application and Research Center (ROMER).

\bibliography{sn-bibliography}

\end{document}